\documentclass[conference]{IEEEtran}
\IEEEoverridecommandlockouts
\usepackage{cite}
\usepackage{amsmath,amssymb,amsfonts}
\usepackage{algorithmic}
\usepackage{graphicx}
\usepackage{textcomp}
\usepackage{xcolor}
\usepackage{hyperref}
\usepackage{multirow}
\usepackage{makecell}
\usepackage{tabularx}
\usepackage[T1]{fontenc}
\usepackage{framed}

\def\BibTeX{{\rm B\kern-.05em{\sc i\kern-.025em b}\kern-.08em
    T\kern-.1667em\lower.7ex\hbox{E}\kern-.125emX}}
\begin{document}

\title{A Deep Learning Framework for Visual Attention Prediction and Analysis of News Interfaces\\
\thanks{This work is part of the project `AI-Driven Media Representation Analysis for Social Equity' (AIRAS). It is financed by Xjenza Malta and The Malta Digital Innovation Authority for and on behalf of the Foundation for Science and Technology through the FUSION: R\&I Research Excellence Programme.}}

\makeatletter
\newcommand{\linebreakand}{%
  \end{@IEEEauthorhalign}
  \hfill\mbox{}\par
  \mbox{}\hfill\begin{@IEEEauthorhalign}
}
\makeatother

\author{
\centering
    \IEEEauthorblockN{Matthew Kenely}
    \IEEEauthorblockA{\textit{Dept. of AI} \\
    \textit{University of Malta}\\
    Msida, Malta \\
    \href{mailto:matthew.kenely@ieee.org}{matthew.kenely@ieee.org}} \\
    \and
    \IEEEauthorblockN{Dylan Seychell}
    \IEEEauthorblockA{\textit{Dept. of AI}  \\
    \textit{University of Malta}  \\
    Msida, Malta \\
    \href{mailto:dylan.seychell@um.edu.mt}{dylan.seychell@um.edu.mt}} \\
    \and
    \IEEEauthorblockN{Carl James Debono}
    \IEEEauthorblockA{\textit{Dept. of Comm. \& Comp. Eng.}  \\
    \textit{University of Malta}  \\
    Msida, Malta \\
    \href{mailto:c.debono@ieee.org}{c.debono@ieee.org}} \\
    \and
    \IEEEauthorblockN{Chris Porter}
    \IEEEauthorblockA{\textit{Dept. of Comp. Info. Systems}  \\
    \textit{University of Malta}  \\
    Msida, Malta \\
    \href{mailto:chris.porter@um.edu.mt}{chris.porter@um.edu.mt}} \\
}

\maketitle

\begin{abstract}
News outlets' competition for attention in news interfaces has highlighted the need for demographically-aware saliency prediction models. Despite recent advancements in saliency detection applied to user interfaces (UI), existing datasets are limited in size and demographic representation. We present a deep learning framework that enhances the SaRa (Saliency Ranking) model with DeepGaze IIE, improving Salient Object Ranking (SOR) performance by 10.7\%. Our framework optimizes three key components: saliency map generation, grid segment scoring, and map normalization. Through a two-fold experiment using eye-tracking (30 participants) and mouse-tracking (375 participants aged 13--70), we analyze attention patterns across demographic groups. Statistical analysis reveals significant age-based variations (p < 0.05, $\epsilon^2$ = 0.042), with older users (36--70) engaging more with textual content and younger users (13--35) interacting more with images. Mouse-tracking data closely approximates eye-tracking behavior (sAUC = 0.86) and identifies UI elements that immediately stand out, validating its use in large-scale studies. We conclude that saliency studies should prioritize gathering data from a larger, demographically representative sample and report exact demographic distributions.
\end{abstract}


\noindent\begin{framed}
\footnotesize
\noindent This is a preprint submitted to the 2025 IEEE Conference on Artificial Intelligence (CAI).
\end{framed}

\section{Introduction}
The growing demand for a human visual system model to predict gaze behavior and ensure accountable, user-friendly user interfaces (UI) in news websites has highlighted the potential of saliency, a subfield of computer vision \cite{seychell2024ai, oparaugo_relevance_2021}.

Although saliency prediction has been applied to user interfaces \cite{gupta_saliency_2018, leiva_understanding_2020, krafka_eye_2016}, existing datasets are smaller than those used to train state-of-the-art models for traditional photographs \cite{jiangSALICONSaliencyContext2015, li2014secrets, wang2017learning}. Moreover, prior studies often feature narrow demographics \cite{shen_webpage_2014, jiangSALICONSaliencyContext2015, juddLearningPredictWhere2009}, small sample sizes \cite{shen_webpage_2014, jiangSALICONSaliencyContext2015, liu2010learning, juddLearningPredictWhere2009}, or fail to report participant demographics in detail \cite{leiva_understanding_2020, jiang2023ueyes, jiangSALICONSaliencyContext2015, liu2010learning, juddLearningPredictWhere2009}. This study underscores the importance of precise demographic reporting in data-driven saliency research.

This study makes three primary contributions:
\begin{enumerate}
    \item The optimization of an existing saliency ranking framework (SaRa) which can generate the ranks of elements in an interface by using any saliency model as a backbone and passing element masks as input.
    \item The curation of a dataset in a typical UI A/B-testing evaluation context which captures attention shifts in news websites. Gaze data was gathered through an eye-tracking experiment (30 participants) and a mouse-tracking experiment (375 participants). The exact demographic distribution of the participants are reported.
    \item Statistical tests are carried out on the responses of the diverse participant base in the mouse-tracking experiment to show any biases affecting visual attention present in the different age and gender groups.
\end{enumerate}

\section{Related Work}
\subsection{Demographic Representation in Saliency Applications}
\label{subsec:saliencyinUI}
Recent applications of automatic saliency detection to UI remain limited. Gupta \textit{et al.} \cite{gupta_saliency_2018} developed a deep learning model for saliency prediction on mobile UI elements, collecting gaze data from 111 participants (aged 19–46) without reporting gender or detailed age distribution. Similarly, Leiva \textit{et al.} \cite{leiva_understanding_2020} analyzed gaze data from 30 participants (average age 25.9) on 193 mobile UIs, though specific age distribution was not provided. Shen \textit{et al.} \cite{shen_webpage_2014} and Jiang \textit{et al.} \cite{jiang2023ueyes} studied 11 and 62 participants respectively, with Jiang reporting gender (23 males, 43 females) but lacking precise age data. These datasets, though valuable, are small and demographically narrow, potentially biasing gaze predictions.

This issue extends beyond UI studies. Widely used saliency datasets (SALICON \cite{jiangSALICONSaliencyContext2015}, MSRA \cite{liu2010learning}, MIT1003 \cite{juddLearningPredictWhere2009}) also rely on small participant groups ($<16$) without reporting age distributions, limiting model generalizability and potentially compromising safety.

\subsection{Mouse-tracking as a Complement to Eye-tracking}
\label{subsec:salicon}
While eye-tracking technology can be considered more sophisticated, mouse-tracking provides complementary insights. Jiang \textit{et al.} \cite{jiangSALICONSaliencyContext2015} demonstrated that mouse-tracking closely approximates eye-tracking, achieving an sAUC of 0.86 compared to 0.89 for eye-tracking, outperforming saliency models (sAUC $< 0.8$). This highlights mouse-tracking's potential for large-scale studies with diverse participants. In this study, we leverage mouse-tracking to collect large-scale data and evaluate its applicability in a UI context.
\section{Saliency Ranking Framework}
\subsection{Original Saliency Ranking Model}
Seychell and Debono \cite{seychell_ranking_2018} introduced SaRa, a framework that segments images and ranks segment saliency using any saliency model as backbone. SaRa divides the input image into a $k \times k$ grid $G$, generating a saliency map where each segment $s$ is scored based on entropy, center bias, and optional depth values:

\begin{equation}
\label{eqn:scoreSegment}
S_{s} = w_H \cdot  H_{s} + w_{CB} \cdot CB_{s} + w_{DS} \cdot DS_{s}
\end{equation}

\subsection{Proposed Saliency Ranking Model}
\subsubsection{Optimization \texorpdfstring{$\sigma$}{Lg} -- Saliency Map Generator}  
We replace the saliency map generator in the original framework proposed by Itti \textit{et al.} \cite{ittiModelSaliencybasedVisual1998} with DeepGaze IIE \cite{linardos_deepgaze_2021}. The MIT/Tuebingen Saliency Benchmark \cite{mit-tuebingen-saliency-benchmark} (Table \ref{tab:mitresults}) shows that DeepGaze IIE outperforms Itti's method and all other benchmarked techniques across all metrics on the MIT300 dataset \cite{judd_benchmark_2012}.

\renewcommand{\arraystretch}{1}
\begin{table}[!htbp]
  \caption{Performance comparison of Itti's model and DeepGaze IIE on MIT/Tuebingen Saliency Benchmark metrics. Metrics with $\downarrow$ indicate that lower values are better.} 
  \label{tab:mitresults}
  \begin{tabular}{lccccccc}
    \hline
    \textbf{Model} & \textbf{IG} & \textbf{AUC} & \textbf{sAUC} & \textbf{NSS} & \textbf{CC} & \textbf{KLDiv $\downarrow$} & \textbf{SIM} \\
    \hline
    Itti \textit{et al.}    &N/A    &0.54   &0.54   &0.41   &0.13   &1.50   &0.34 \\
    DG IIE            &1.07   &\textbf{0.88}   &\textbf{0.79}   &\textbf{2.53}   &\textbf{0.82}   &\textbf{0.35}   &\textbf{0.70} \\
    \hline
  \end{tabular}
\end{table}  

\subsubsection{Optimization \texorpdfstring{$\epsilon$}{Lg} -- Grid Segment Saliency Score Equation}  
In order to better leverage DeepGaze IIE's capabilities, we assign greater weight to the direct pixel values (labeled $SS_{s}$) in the generated saliency map. A revised saliency score formula is proposed. For the purpose of this study, score components weights are set to 1.

\begin{equation}
\label{eqn:scoreSegmentWithSum}
S_{s} = w_H \cdot H_{s} + w_{SS} \cdot SS_{s} + w_{CB} \cdot CB_{s} + w_{DS} \cdot DS_{s}
\end{equation}

\subsection{Optimization \texorpdfstring{$\nu$}{Lg} -- Saliency Map Normalization}  
Saliency maps may contain noise in non-salient regions that can inflate entropy values, particularly with high bit depths. We include a post-processing step in the pipeline which applies a $31\times31$ Gaussian filter, normalizes to $[0, 255]$, and reduces bit depth from $2^8$ to $2^5$ by dividing values by 8. This approach minimises noise and results in 32 discrete saliency levels. Hyperparameters were optimized on a 482-image subset of MS-COCO (Figure \ref{fig:noblurvsblurdeepgaze}).

\begin{figure}[!htbp]
    \centering
    \includegraphics[width=0.45\textwidth]{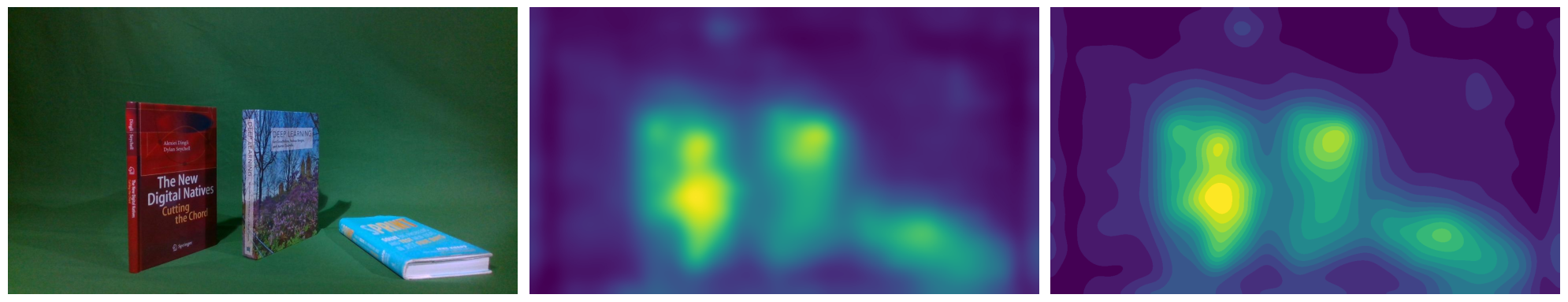}
    \caption[Normalization optimization comparison using DeepGaze IIE.]{From left to right: Original image, saliency map from DeepGaze IIE, and map after $31\times31$ Gaussian filter and bit depth normalization.}
    \label{fig:noblurvsblurdeepgaze}
\end{figure}  

\subsection{Evaluation}
\label{subsec:saraevaluation}  
Spearman's rank correlation coefficient (SRCC) measures the strength of a monotonic relationship between two variables \cite{sedgwick2014spearman}. It is well-suited for saliency ranking tasks and frequently used in related work \cite{wang2001comparative, kalash2019relative, niu2018meta}. Unlike linear correlation, SRCC detects correlations in relative rank order.  

SRCC returns $\rho \in [-1, 1]$, where 1 indicates perfect rank order, -1 indicates complete contrast and $\rho$ is calculated as:

\begin{equation}
\label{eq:spearman}
\rho = 1 - \frac{6 \sum d_i^2}{n(n^2 - 1)}
\end{equation}

where $\sum d_i^2$ is the sum of squared rank differences, and $n$ is the sample size.  

The Salient Object Ranking (SOR) metric, introduced by Islam \textit{et al.} \cite{islamRevisitingSalientObject2018}, normalizes SRCC to $[0, 1]$ for clearer interpretation. The optimized SaRa framework will be quantitatively evaluated for saliency ranking on a dataset combining MS-COCO object masks and SALICON fixation sequences using SOR, following the approach in \cite{sirisInferringAttentionShift2020}.

\subsection{Discussion}
\begin{table}[!htbp]
\renewcommand{\arraystretch}{1}
\caption{Results of the quantitative experiment in \cite{sirisInferringAttentionShift2020} comparing state-of-the-art saliency models (average weighted by images used) and SaRa with combinations of optimizations in order performance improvement magnitude.}
\label{tab:quantitativeresults}

\centering
\begin{tabular}{lcc}
\hline
\textbf{Model}                                       & \textbf{SOR $\uparrow$}       & \textbf{\#Images used $\uparrow$} \\
\hline
RSDNet                & 0.728                & 2418                     \\
S4Net                          & 0.891                & 1507                     \\
BASNet                         & 0.707                & 2402                     \\
CPD-R                           & 0.766                & 2417                     \\
SCRN    & 0.756                & 2418                     \\
Siris \textit{et al.} & 0.792           & 2365                     \\
\textit{Average}                                & 0.765       & 2278            \\
\hline
Original SaRa \cite{seychell_ranking_2018} & 0.654  & 2347                     \\
SaRa + $\nu$                         & 0.670       & 2347           \\
SaRa + $\epsilon$                        & 0.685       & 2347           \\
SaRa + $\sigma$                         & 0.714       & 2347           \\
SaRa + $\epsilon\sigma$                   & 0.715       & 2347           \\
SaRa + $\epsilon\sigma\nu$           & 0.718       & 2347           \\
SaRa + $\epsilon\sigma\nu$, $k = 30$           & 0.724       & 2347           \\
\hline
\end{tabular}
\end{table}

\label{subsec:quantitativediscussion}
Table \ref{tab:quantitativeresults} demonstrates that each optimization incrementally improved SOR performance. Notably, the Grid Segment Saliency Score Equation enhanced performance even when using Itti's model, while DeepGaze IIE provided the most substantial boost. A segment grid size of $k = 30$ balanced performance with computational efficiency. ($\mathcal{O}(n^2)$ complexity). Applying all optimizations achieved a 10.7\% SOR increase over the original technique, reaching performance comparable to state-of-the-art models
\section{Methodology}
\subsection{Gaze Dataset}
\label{sec:gazedataset}
The dataset comprises 10 pairs of news website interfaces in desktop and mobile forms. Each pair includes a control version with distracting elements and an experimental version with these elements removed. Differences between versions, termed Areas of Interest (AOIs), highlight the impact of distractions on gaze attention. Data and implementation available on GitHub\footnote{\href{https://github.com/matthewkenely/framework-attention-news}{https://github.com/matthewkenely/framework-attention-news}}.

\subsection{Eye-tracking Experiment}
\label{sec:experimentA}
This experiment established a baseline for comparing gaze and mouse-tracking behaviors. Thirty participants were split into control and experimental groups, viewing interfaces with either highly salient or neutral elements. Gaze data was recorded using a GazePoint eye-tracker at 60 Hz, with 9-point calibration for accuracy.

Participants viewed 10 interfaces for 10 seconds each, in random order to minimize exposure bias. Followed by a questionnaire on demographics (age, gender, or ``rather not say'') and awareness of distracting elements. While only 5 participants were female, statistical analysis showed minimal gender influence on gaze patterns. Ages ranged from 19 to 26, potentially biasing results and emphasizing the need for a more age-diverse dataset (see Table \ref{tab:kruskalwallistest}).

\subsection{Mouse-tracking Experiment}
\label{sec:experimentB}
This online experiment engaged 375 participants, offering insights into demographic influences on attention. Participants viewed 10 news interfaces, interacting by hovering or clicking on elements of interest. Mouse movements and clicks/taps were tracked, generating attention heatmaps.

\begin{figure}[!htbp]
    \centering
    \includegraphics[width=0.33\textwidth,keepaspectratio]{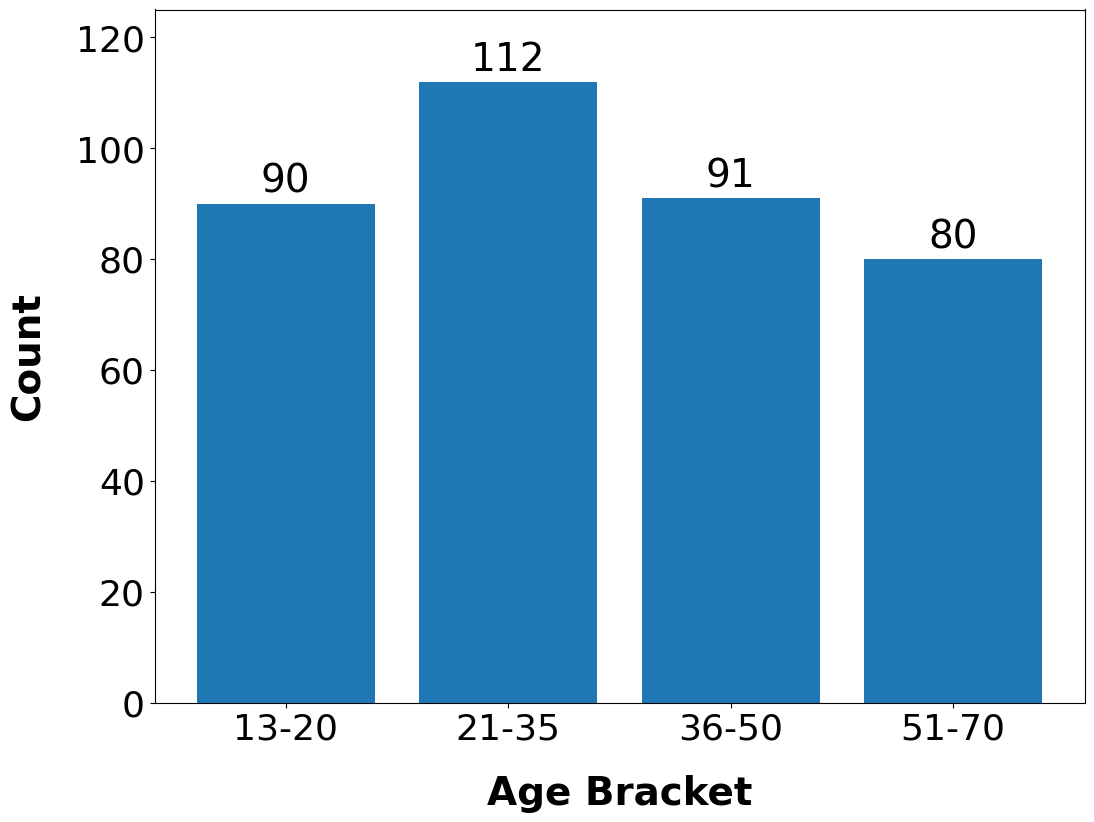}
    \caption[Age distribution in the mouse-tracking experiment.]{Age distribution in the mouse-tracking experiment, binned into 4 groups.}
    \label{fig:agedistribution}
\end{figure}

Participants provided their age and gender (or ``Rather not say'') before assignment to control or experimental groups. Each group viewed a unique shuffled sequence to mitigate bias. Desktop users hovered over a central dot before each image for standardization. Mouse data was tracked using JavaScript's MouseEvent API and stored in JSON on Firebase Cloud.

Among participants, 64\% were female (131 male, 240 female, 3 other, 1 rather not say), creating a gender imbalance. However, a Mann-Whitney U test confirmed that gender had minimal influence on attention. The evenly distributed age groups (Figure \ref{fig:agedistribution}) enabled Kruskal-Wallis tests to examine age-attention correlations, informing further qualitative heatmap analyses. Levene's Test was used to ensure variance equality.

The following are reported in Table \ref{tab:kruskalwallistest}:
\begin{itemize}
    \item p-Value (\textit{p}): Indicates the probability of observing the test statistic under the null hypothesis (that there is no significant difference in the click/tap location based on the demographic variable). A strong likelihood of statistical significance is assumed at p $< 0.05$.
    \item Effect Size (\(\varepsilon^2\)): Measures the variance explained by the grouping variable, indicating practical significance.
\end{itemize}

\section{Evaluation}
\subsection{Dataset}
\begin{table*}[htbp]
\caption[Results from the eye-tracking experiment compared to rank shifts detected by SaRa.]{Results from the eye-tracking experiment compared to rank shifts detected by SaRa. Each result concerns the AOI(s) in the interfaces. Better performance, denoted in bold, implies that the interface was less distracting to participants. TMI refers to The Malta Independent.}
\label{tab:fixationstats}

\renewcommand{\arraystretch}{1}
\begin{tabularx}{\textwidth}{Xcccccccccccc}
\hline
\multirow{2}{*}{\textbf{Image}} & \multicolumn{2}{c}{\makecell{\textbf{Time}\\\textbf{Viewed \% ↓}}} & \multicolumn{2}{c}{\makecell{\textbf{Avg.}\\\textbf{Fixations ↓}}} & \multicolumn{2}{c}{\makecell{\textbf{Revisitors\% ↓}}} & \multicolumn{2}{c}{\makecell{\textbf{Avg.}\\\textbf{Revisits ↓}}} & \multicolumn{2}{c}{\makecell{\textbf{Avg. 1st}\\\textbf{View ↑}}} & \multicolumn{2}{c}{\makecell{\textbf{SaRa}\\\textbf{Rank ↑}}} \\
                       & CTRL              & EXPR             & CTRL              & EXPR             & CTRL             & EXPR             & CTRL             & EXPR             & CTRL                 & EXPR                 & CTRL           & EXPR           \\

\hline
Custom (DESKTOP)       & 17.90             & \textbf{14.03}   & 7.73              & \textbf{6.00}    & 93.33            & \textbf{86.67}   & 3.43             & \textbf{2.69}    & 0.57                 & \textbf{1.13}        & 1.00           & \textbf{5.00}  \\
Custom (MOBILE)        & 7.60              & \textbf{2.56}    & 3.13              & \textbf{2.00}    & 73.33            & \textbf{13.33}   & \textbf{2.27}    & 3.50             & 2.27                 & \textbf{6.40}        & 1.00           & \textbf{2.00}  \\
Times of Malta (1)     & \textbf{28.16}    & 32.60            & \textbf{10.60}    & 10.73            & 86.67            & 86.67            & 2.77             & \textbf{1.69}    & 0.37                 & \textbf{0.50}        & 1.00           & 1.00           \\
Lovin' Malta           & 30.91             & \textbf{5.62}    & 12.07             & \textbf{2.44}    & 100.00           & \textbf{13.33}   & 4.27             & \textbf{1.25}    & 0.84                 & \textbf{1.59}        & 1.00           & \textbf{4.00}  \\
Illum                  & 8.04              & \textbf{7.78}    & 4.83              & \textbf{3.64}    & 80.00            & 80.00            & 3.08             & \textbf{2.58}    & 1.44                 & \textbf{2.18}        & 4.00           & \textbf{6.00}  \\
TMI  & 6.59              & \textbf{5.65}    & 3.04              & \textbf{2.96}    & \textbf{16.67}   & 20.00            & \textbf{2.50}    & 2.60             & 2.65                 & \textbf{4.17}        & \textbf{6.00}  & 2.00           \\
Malta Today            & 10.87             & \textbf{5.67}    & 5.28              & \textbf{2.50}    & 43.33            & \textbf{40.00}   & 3.48             & \textbf{2.33}    & \textbf{4.61}        & 3.52                 & 2.00           & \textbf{4.00}  \\
The Shift              & 11.28             & 4.76             & 5.17              & \textbf{1.75}    & 20.00            & \textbf{0.00}    & 3.20             & \textbf{0.00}    & 5.24                 & \textbf{8.89}        & 3.00           & \textbf{4.00}  \\
Times of Malta (2)     & \textbf{10.89}    & 13.13            & 5.05              & \textbf{5.01}    & 33.33            & \textbf{16.67}   & 3.14             & \textbf{2.33}    & \textbf{4.73}        & 4.12                 & 2.00           & \textbf{7.00}  \\
TVM                    & 12.06             & \textbf{10.66}   & 5.08              & 4.07             & 73.33            & \textbf{46.67}   & 2.00             & \textbf{1.71}    & 2.61                 & \textbf{3.18}        & 5.00           & \textbf{7.00}  \\
\hline
\textit{Average}                & 14.43             & \textbf{10.25}   & 5.08              & \textbf{4.12}    & 62.00            & \textbf{40.33}   & 3.01             & \textbf{2.07}    & 2.53                 & \textbf{3.57}        & 2.60           & \textbf{4.20} \\
\hline
\end{tabularx}
\end{table*}

\begin{figure}[!htbp]
    \centering
    \includegraphics[width=0.33\textwidth,keepaspectratio]{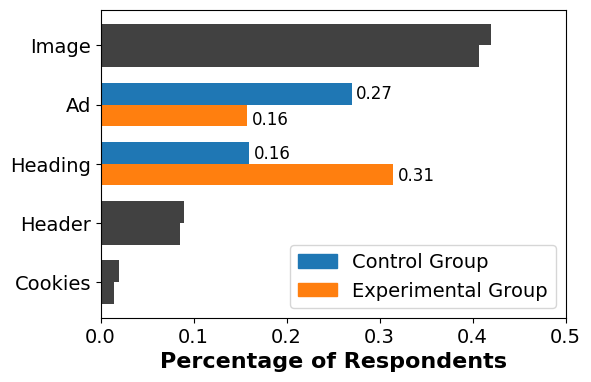}
    \caption[Responses to question 2 from the eye-tracking experiment.]{Responses to ``Which type of element do you feel stood out the most?'' from the control group (blue) and the experimental group (orange).}
    \label{fig:q2}
\end{figure}

We observe an apparent reduction in the distraction factor of the AOIs in most interfaces shown to the experimental group. As shown in Table \ref{tab:fixationstats}, GazePoint data reveals that, on average, AOIs were viewed 4.2\% less (0.42s), fixated on 0.96 fewer times, revisited by 21.7\% fewer participants, revisited 0.94 times less, and first viewed 1.03 seconds later.

The eye-tracking questionnaire further supports this, with Figure \ref{fig:q2} showing a significant shift in attention. Participants in the experimental group focused more on article headings (relevant content), while attention to images dropped by 0.3\% and to advertisements by 10.3\%.

\subsection{Eye-tracking vs Mouse-tracking}
\begin{figure}[!htbp]
    \centering
    \includegraphics[width=0.3\textwidth,keepaspectratio]{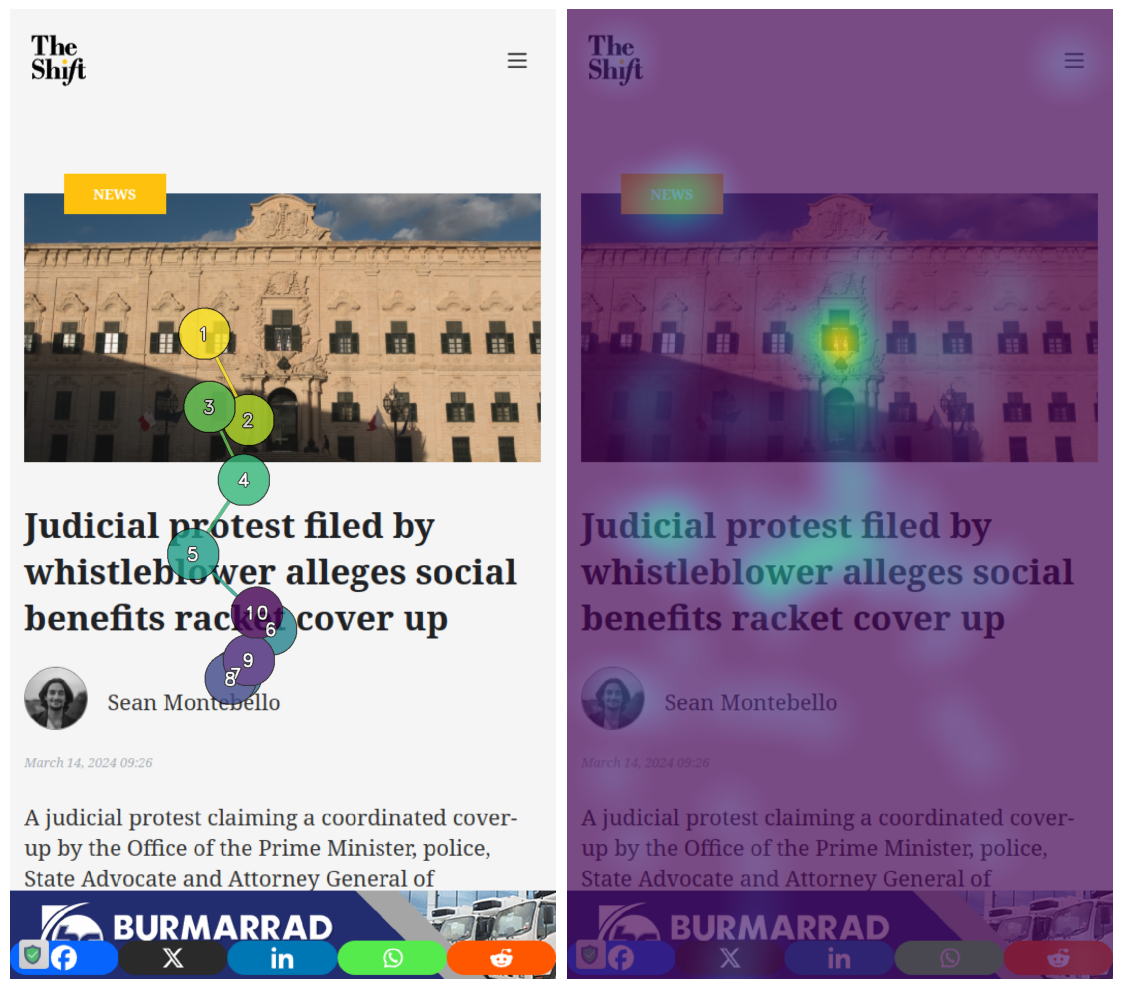}
    \caption[``The Shift'' eye-tracking vs mouse-tracking comparison.]{Gaze location results for the interface ``The Shift'' shown to the experimental group. Left: average fixation location per second in the eye-tracking experiment, right: heatmap from the mouse-tracking experiment.}
    \label{fig:eyevsmouse2}
\end{figure}
\label{subsec:qualitativediscussion}
Heatmaps reveal that eye-tracking highlights elements that sustain attention over time (e.g., 10 seconds), while mouse-tracking captures what initially stands out on the interface. In this UI context, eye-tracking participants tended to read through screen captures, as shown in Figure \ref{fig:eyevsmouse2}.

Notably, early fixation locations from eye-tracking align with the top salient regions that are identified by mouse-tracking, highlighting their complementary roles. Eye-tracking suits models for sustained attention, while mouse-tracking better reflects initial visual saliency. Expanding on the work done by Jiang et al. in Subsection \ref{subsec:salicon}, we suggest that mouse-tracking should augment, rather than replace, eye-tracking in UI studies.

\subsection{Demographic Findings}
\label{subsec:demographicfindings}

\renewcommand{\arraystretch}{1}
\begin{table}[!t]
  \caption{Results of the age Kruskal‐Wallis Test. Null hypothesis rejections underlined. X: horizontal gaze movement. Y: vertical gaze movement.}
  \label{tab:kruskalwallistest}
\begin{tabular}{>{\raggedright\arraybackslash}p{0.14\textwidth} >{\centering\arraybackslash}p{0.04\textwidth} >{\centering\arraybackslash}p{0.035\textwidth} >{\centering\arraybackslash}p{0.045\textwidth} >{\centering\arraybackslash}p{0.035\textwidth} >{\centering\arraybackslash}p{0.045\textwidth}}
\hline
\textbf{Image}                 & \textbf{Group}       & \textit{\textbf{p (X)}} & \textit{\textbf{$\varepsilon^2$ (X)}} & \textit{\textbf{p (Y)}} & \textit{\textbf{$\varepsilon^2$ (Y)}} \\
\hline
Custom (DESKTOP)               & CTRL                 & 0.416               & 0.000 & 0.155               & 0.013 \\
\textbf{Custom (MOBILE)}       & \textbf{CTRL}        & 0.137       & 0.014 & \underline{\textbf{0.038}} & \textbf{0.031} \\
Times of Malta 1               & CTRL                 & 0.686               & 0.000 & 0.710               & 0.000 \\
Lovin' Malta                   & CTRL                 & 0.123               & 0.016 & 0.209               & 0.009 \\
\textbf{Illum}                 & \textbf{CTRL}        & \underline{\textbf{0.042}} & \textbf{0.029} & 0.792       & 0.000 \\
TMI          & CTRL                 & 0.395               & 0.000 & 0.317               & 0.003 \\
\textbf{Malta Today}           & \textbf{CTRL}        & 0.549       & 0.000 & \underline{\textbf{0.019}} & \textbf{0.039} \\
The Shift                      & CTRL                 & 0.844               & 0.000 & 0.684               & 0.000 \\
Times of Malta 2               & CTRL                 & 0.704               & 0.000 & 0.953               & 0.000 \\
TVM                            & CTRL                 & 0.165               & 0.012 & 0.510               & 0.000 \\
Custom (DESKTOP)               & EXPR                 & 0.480               & 0.000 & 0.404               & 0.000 \\
Custom (MOBILE)                & EXPR                 & 0.128               & 0.000 & 0.762               & 0.000 \\
Times of Malta 1               & EXPR                 & 0.477               & 0.015 & 0.206               & 0.009 \\
Lovin' Malta                   & EXPR                 & 0.683               & 0.000 & 0.350               & 0.000 \\
\textbf{Illum}                 & \textbf{EXPR}        & 0.929       & 0.000 & \underline{\textbf{0.015}} & \textbf{0.042} \\
TMI          & EXPR                 & 0.985               & 0.009 & 0.213               & 0.008 \\
Malta Today                    & EXPR                 & 0.151               & 0.000 & 0.146               & 0.014 \\
\textbf{The Shift}             & \textbf{EXPR}        & \underline{\textbf{0.019}} & \textbf{0.040} & 0.532       & 0.000 \\
Times of Malta 2               & EXPR                 & 0.543               & 0.000 & 0.356               & 0.001 \\
TVM                            & EXPR                 & 0.620               & 0.000 & 0.619               & 0.000 \\
\hline
\end{tabular}
\end{table}

\begin{figure*}[!htbp]
    \centering
    \includegraphics[width=0.85\textwidth,keepaspectratio]{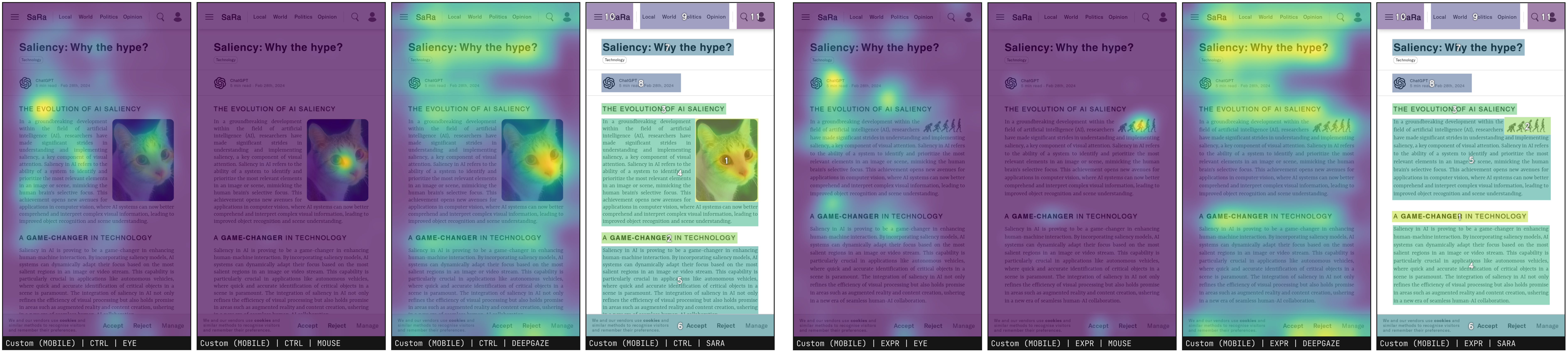}
    \caption[Custom (MOBILE) interface heatmaps and SaRa ranks.]{Custom (MOBILE) interface heatmaps. Control group on the left, experimental group at the right. From left to right for each group: heatmaps from the eye-tracking experiment, heatmaps from the mouse-tracking experiment, saliency maps generated by DeepGaze IIE and the corresponding SaRa ranks.}
    \label{fig:summary1}
\end{figure*}

\begin{figure*}[!htbp]
    \centering
    \includegraphics[width=0.85\textwidth,keepaspectratio]{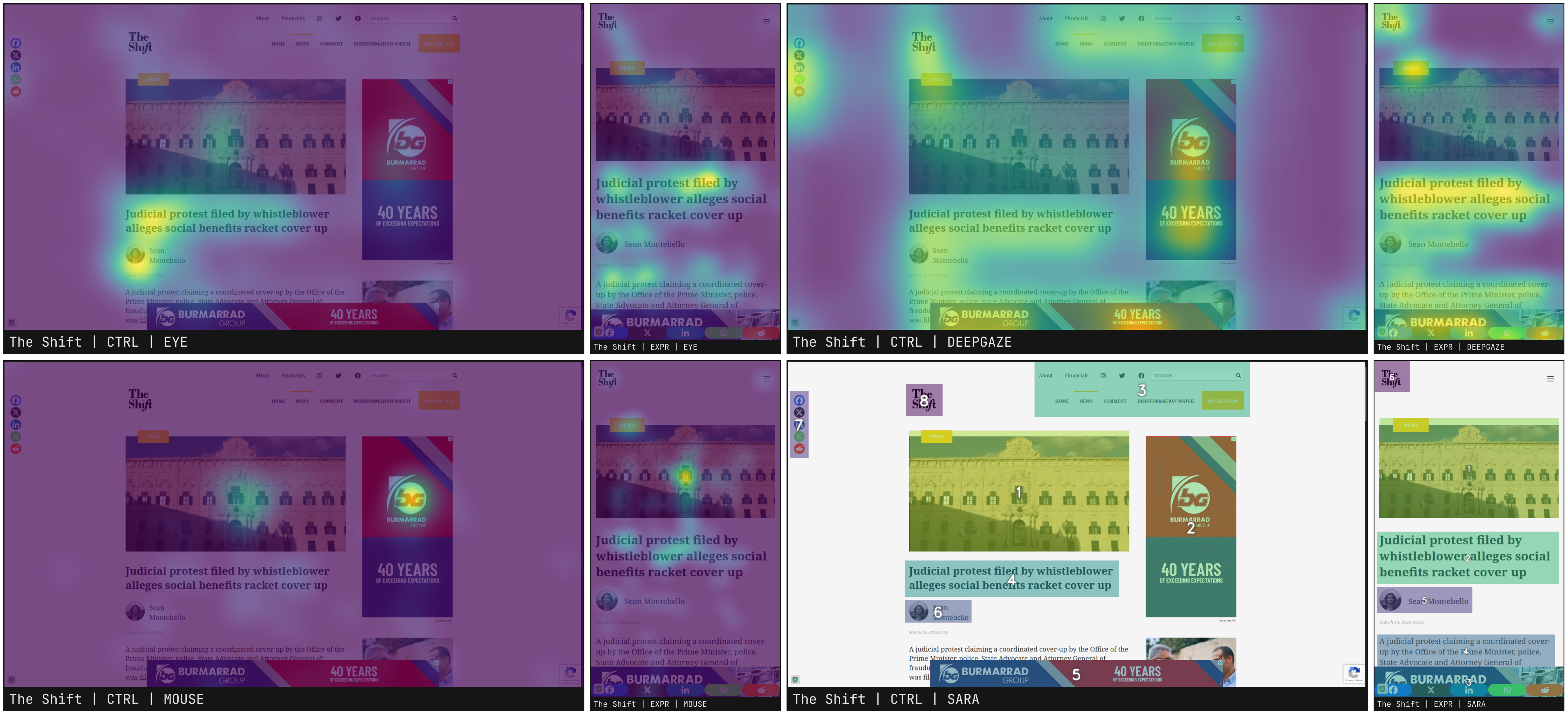}
    \caption[The Shift interface heatmaps and SaRa ranks.]{The Shift interface heatmaps. Within each pair, the control group is on the left and the experimental group is on the right. Top-left: heatmaps from the eye-tracking experiment; bottom-left: heatmaps from the mouse-tracking experiment; top-right: saliency maps generated by DeepGaze IIE; bottom-right: the corresponding SaRa ranks.}
    \label{fig:summary3}
\end{figure*}

The demographic statistical tests and subsequent qualitative analyses reveal notable differences in gaze tendencies across demographic groups.

Gender played a weaker role in influencing gaze patterns. The only significant result from the Mann-Whitney U Test occurred with the ``The Malta Independent'' interface shown to the control group. A bias toward the first article image was evident among female respondents, potentially due to their superior ability to recognize faces \cite{herlitz2013sex, proverbio2017sex}.

Age, on the other hand, was shown to have a much stronger influence on where people were likely to look, with the difference between the gaze tendencies of the age groups being statistically significant in 5 out of the 20 examined interfaces (25\%). We observed the following through heatmap analyses:

\begin{enumerate}
    \item Within headings, the specific words which stood out to participants tended to shift based on their age group, e.g. ``Judicial'' and ``whistleblower'' in the 36--50 age group and ``protest'' in the 51--70 age group.
    \item Participants in the 21--35 age group were more likely to reject cookies, whereas the 36--70 age group were more likely to accept them;
    \item Older demographics (36--70) were more likely to look at news article headings rather than the image;
    \item Participants tended to look at images featuring people who are the same age as them;
\end{enumerate}

\subsection{Saliency Ranking Framework}
This section discusses the findings from both experiments in comparison to the predictions of the AI framework. We present interfaces from two shift types -- content and responsiveness -- where discrepancies were found between demographic groups. The potential effects of the narrow demographic range in the eye-tracking experiment (mostly male, ages 19--26) are cross-checked with the demographic findings in Subsection \ref{subsec:demographicfindings}.

\subsubsection{Custom (Mobile)}
The results and SaRa ranks for this interface are shown in Figure \ref{fig:summary1}. This custom interface aimed to assess content attention shifts by including and removing the primary cat image as the AOI. The shift was less significant, with the image still receiving considerable attention. DeepGaze IIE and the resulting SaRa ranks accurately captured this, with the ablated AOI receiving a rank difference of only $-1$, and the attention shift toward the second header, ``A Game-Changer in Technology'', was also well represented. Demographic heatmap analyses revealed that younger participants were more likely to direct attention to the ``Accept'' button in the cookies bar. This biased behavior is apparent in the eye-tracking experiment (ages 19--26).

\subsubsection{The Shift}
The results and SaRa ranks for this interface are shown in Figure \ref{fig:summary3}. This interface aimed to assess responsiveness attention shifts by comparing desktop and mobile versions, specifically the ablation of the large ad on the right. In both experiments, attention toward the ad from the experimental group was negligible, with the focus shifting to the main image, heading, and content. DeepGaze IIE and the corresponding SaRa ranks captured this shift well, with the AOIs receiving a rank shift of $-1$. However, the top bar in the control group, which received no attention, was erroneously assigned rank 2 due to entropy. Demographic heatmap analyses revealed that younger participants were more likely to show attention to the article image. Again, this biased behavior is apparent in the eye-tracking experiment.
```
\section{Discussion}
\subsection{Demographic Findings}
Past research has often focused on young adults (21–35), potentially overlooking key differences in attention patterns across younger and older demographics. Future studies should prioritize both participant quantity and diversity. A large sample size alone does not guarantee representativeness if it does not account for demographic variations such as age, gender, and digital literacy.

For AI-based UI evaluation tools to be effective for a broad audience, training data must reflect user diversity. Models trained mainly on younger, tech-savvy participants \cite{dingli2015new} may exhibit biases, neglecting the preferences and limitations of older or less tech-savvy users \cite{alkali2004experiments}. Participant recruitment should be methodical, using stratified sampling to ensure accurate representation of diverse groups based on regional and national demographic data. Depending on the application, this may involve narrowing or broadening the participant base. General-purpose interfaces require diverse representation to ensure inclusivity.

\subsection{Experiments and the Ranking Framework}
The AI framework compose of DeepGaze IIE and SaRa performed well in capturing attention across control and experimental groups in various interface designs. As per Table \ref{tab:fixationstats}, it excelled at predicting attention shifts, especially regarding AOIs. However, DeepGaze IIE struggled with interpreting semantic meaning in images and capturing saliency for distracting images, which was expected since it was trained on traditional photographs rather than user interfaces. This limitation was mitigated by the entropy component in Equation (\ref{eqn:scoreSegmentWithSum}), which favors large elements.

\section{Conclusion}%
\label{sec:conclusion}
This study proposed a demographically representative dataset to capture attention shifts in responsive interfaces when distracting elements are ablated, or the structure is altered. This dataset, from eye-tracking and mouse-tracking experiments, was used to evaluate the SaRa saliency ranking framework, which was optimized for better performance in the SOR metric. SaRa effectively assesses the fairness of attention distribution in user interfaces through saliency prediction.

Our two-fold experimentation revealed that eye-tracking and mouse-tracking capture different aspects of attention: eye-tracking focuses on sustained attention, while mouse-tracking captures immediate attention. This distinction is important when training saliency models for UI evaluation, as the data collection method influences the patterns models learn.

The demographic analysis highlights the need for representative participant bases when curating datasets for saliency model training. Age, in particular, influences attention patterns. A transparent approach to dataset curation will ensure the generalizability of AI-based UI evaluation tools and encourage user-centric UI.

\bibliographystyle{IEEEtran}
\bibliography{references}

\end{document}